\documentclass[final]{cvpr}

\usepackage{times}
\usepackage{epsfig}
\usepackage{graphicx}
\usepackage{amsmath}
\usepackage{amssymb}

\usepackage{url}            
\usepackage{booktabs}       
\usepackage{amsfonts}       
\usepackage{nicefrac}       
\usepackage{microtype}      

\usepackage{graphicx}
\usepackage{amsmath,amssymb} 
\usepackage{bbold}
\usepackage{dsfont}
\usepackage{enumerate}
\usepackage[toc,page]{appendix}
\usepackage[font=small,labelfont=bf]{caption}

\usepackage{enumitem}

\usepackage{xcolor}
\usepackage{caption}
\usepackage{subcaption}
\usepackage{bm}
\usepackage{isomath}
\usepackage{arydshln}
\usepackage{stmaryrd}

\usepackage{fixltx2e}
\usepackage{dblfloatfix}
\usepackage{pbox}
\usepackage{capt-of}

\usepackage[normalem]{ulem}
\usepackage{multirow}

\usepackage{colortbl}

\usepackage{mathtools}
\usepackage{array}

\makeatletter
\@namedef{ver@everyshi.sty}{}
\makeatother
\usepackage{pgf}

\newcommand{\xpt}{\edef\f@size{\@xpt}\rm}

\def\ie{\emph{i.e.}}
\def\etc{\emph{etc}}

\usepackage{tikz}

\newcommand{\comment}[1]{}

\usepackage{indentfirst}
\renewcommand\vec[1]{\ensuremath\boldsymbol{#1}}
\renewcommand\cdots{...}

\newcommand{\tH}{\vec{\mathcal{H}}}

\newcommand{\tF}{\vec{\mathcal{F}}}

\newcommand{\mY}{\mathbf{Y}}

\newcommand{\mX}{\mathbf{X}}

\newcommand{\mbr}[1]{\mathbb{R}^{#1}}

\newcommand{\idx}[1]{\mathcal{I}_{#1}}

\newcommand{\tR}{\vec{\mathcal{R}}}

\newcommand{\vpsi}{\boldsymbol{\psi}}

\def\eg{\emph{e.g.}}

\newcommand{\tG}{\boldsymbol{\mathcal{G}}}

\newcommand{\mPhi}{\boldsymbol{\Phi}}

\newcommand{\stkout}[1]{{\ifmmode\text{\sout{\ensuremath{#1}}}\else\sout{#1}\fi}}

\usepackage[pagebackref=true,breaklinks=true,colorlinks,bookmarks=false]{hyperref}

\def\cvprPaperID{8703} 
\def\confYear{CVPR 2021}

\begin{document}

\title{Rethinking Class Relations: Absolute-relative Supervised and Unsupervised Few-shot Learning}

\author{Hongguang Zhang\textsuperscript{1,2}\thanks{This work is accepted to CVPR'21.} \qquad Piotr Koniusz\textsuperscript{3,2} \qquad Songlei Jian\textsuperscript{5} \qquad Hongdong Li\textsuperscript{2} \qquad Philip H. S. Torr\textsuperscript{4}\\
$^1$Systems Engineering Institute, AMS \quad $^2$Australian National University \quad$^3$Data61/CSIRO \\
$^4$University of Oxford \quad $^5$National University of Defense Technology \\
{\tt\small firstname.lastname@\{anu.edu.au\textsuperscript{2}, data61.csiro.au\textsuperscript{3},eng.ox.ac.uk\textsuperscript{4}\}}
}

\maketitle

\begin{abstract}
The majority of existing few-shot learning methods describe image relations with binary labels. However, such binary relations are insufficient to teach the network complicated real-world relations, due to the lack of decision smoothness. Furthermore, current few-shot learning models capture only the similarity via relation labels, but they are not exposed to class concepts associated with objects, which is likely detrimental to the classification performance due to underutilization of the available class labels. For instance,  children learn the concept of {\em tiger} from a few of actual examples as well as from comparisons of {\em tiger} to other animals. Thus, we hypothesize that  both similarity and class concept learning must be occurring simultaneously. With these observations at hand, we study the fundamental problem of simplistic class modeling in current few-shot learning methods. We rethink the relations between class concepts, and propose a novel Absolute-relative Learning paradigm to fully take advantage of label information to refine the image an relation representations in both supervised and unsupervised scenarios. Our proposed paradigm improves the performance of several  state-of-the-art models on publicly available datasets. 
\end{abstract}

\section{Introduction}
\label{sec:intro}
Deep learning, a popular learning paradigm in computer vision, has improved the performance on numerous computer vision tasks, such as category recognition, scene understanding and action recognition.  However, deep models  heavily rely on large amounts of labeled training data,  costly data collection and labelling. 

\begin{figure}[t]
    \centering
    \includegraphics[width=0.85\linewidth]{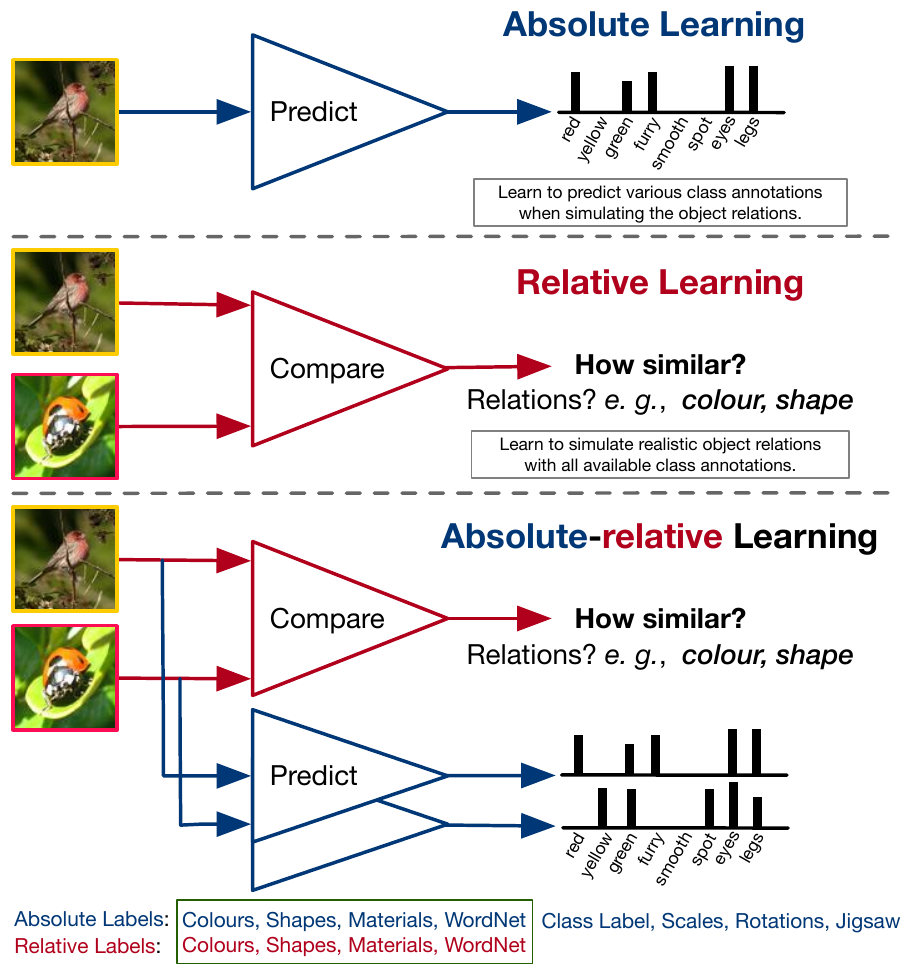}
    \vspace{-0.3cm}
    \caption{\small Our few-shot learning paradigm. Absolute Learning (AL) refers to the strategy where a pipeline learns to predict absolute object information \eg, object or concept class. Relative Learning (RL) denotes similarity (relation) learning with the use of binary $\{0,1\}$ and/or soft $[0;1]$ similarity labels. Absolute-relative Learning (ArL) is a combination of AL and RL, which is akin to multi-task learning, and unary and pair-wise potentials in semantic segmentation. ArL is also  conceptually closer to how humans learn from few examples.}
    \label{fig:arl_demo}
    \vspace{-0.5cm}
\end{figure}

In contrast, humans enjoy the ability to learn and memorize new complex visual concepts from very few examples. Inspired by this observation, researchers have focused on the so-called Few-shot Learning (FSL), for which a network is trained by the use of only few labeled training instances. Recently,  deep networks based on relation-learning have gained the popularity \cite{vinyals2016matching,snell2017prototypical,sung2017learning,NIPS2017_7082,sosn,salnet,zhang2020few,Zhang_2020_CVPR,ziko2020laplacian,deeper_look2,Simon_2020_CVPR}. Such approaches often apply a form of metric learning adapted to the few-shot learning task. They learn object relations (similarity learning on query and support images) based on support classes, and can be evaluated on images containing novel classes. 

However, there are two major problems in these relation learning pipelines, namely,  (i) binary $\{0,1\}$ labels are used to express the similarity between pairs of images, which cannot capture the similarity nuisances in the real-world setting due to the hardness of such modeling, which leads to biases in the relation-based models, (ii) only pair-wise relation labels are used in these pipelines, so the models have no knowledge of the actual class concepts. In other words, these models are trained to learn the similarity between image pairs while they discard the explicit object classes despite they are  accessible in the training stage.

We conjuncture that  these two problems pose inconsistency between current few-shot  learning approaches and human's cognitive processes.  To this end, we propose the Absolute-relative Learning (ArL) which exposes few-shot learners to both similarity and class labels, and we employ semantic annotations to circumvent the issue with the somewhat rigid binary similarity labels $\{0,1\}$.

Our ArL consists of two separate learning modules, namely, Absolute Learning (AL) and Relative Learning (RL). AL denotes the strategy in which we learn to predict the actual object categories or class concepts in addition to learning the class relations. 
In this way, the feature extracting network is exposed to additional object- or concept-related knowledge. RL refers to the similarity learning strategy for which (apart of binary $\{0,1\}$ labels) we employ semantic annotations to promote the realistic similarity between image pairs. We use attributes or  word2vec to obtain the semantic relation labels and learn element-wise similarities \eg, if two objects have same colour, texture, \etc. Such labels are further used as the supervisory cue in relation learning to capture the realistic soft relations between objects beyond the binary similarity. 

By combing AL and RL which constitute on ArL, the relation network is simultaneously taught the class/object concepts together with more realistic class/object relations, thus naturally yielding an improved accuracy. Moreover, we  use the predictions from the absolute and relative learners as interpretable features to promote the original relation learning via feedback connections.

Our approach is somewhat related to multi-modal learning which leverages multiple sources of data for training and testing. However, while multi-modal learning combines multiple streams of data on network inputs, our ArL models the semantic annotations in the label space, that is, we use them as the network output. We believe that  using multiple abstractions of labels (relative \vs absolute) encourages the network to preserve more information about objects relevant to the few-shot learning task. Our strategy  benefits from multi-task learning where two tasks learnt simultaneously help each other to outperform  a naive fusion of two separate tasks. These tasks somewhat resemble  unary and pair-wise potentials in semantic segmentation.

We note that obtaining the semantic information for novel classes (the testing step in few-shot learning) is not always easy or possible. Since our pipeline design is akin to multi-task rather than multi-modal learning, our model does not require additional labeling at the testing stage. Therefore, it is a more realistic setting than that of existing approaches.

In addition to the classic supervised few-shot recognition, we  extend our ArL to the unsupervised scenario. Different with approach \cite{gidaris2019boosting} that merely applies the self-supervised discriminator as an auxiliary task to improve the performance of supervised FSL, we  develop an effective unsupervised FSL based on ArL. As there is no annotations for training samples, we rely on augmentation labelling (\eg, rotations, flips and colors) to perform Absolute-relative Learning. 
Below, we summarize our contributions:
\renewcommand{\labelenumi}{\roman{enumi}.}
\hspace{-1.0cm}
\begin{enumerate}[leftmargin=0.6cm]
    \item We propose so-called Absolute-relative Learning which can be embedded into popular few-shot pipelines to exploit both similarity and object/concept labelling. 
    \item We extend our approach to unsupervised FSL, and we show how to create self-supervised annotations for unsupervised Absolute-relative Learning. 
    \item  We investigate the influence of different types of similarity measures on attributes in Relative Learning to simulate realistic object relations.
    \item We investigate the influence of different Absolute Learning branches on the classification performance.
\end{enumerate}

To the best of our knowledge, we are the first to perform an in-depth analysis of object and class relation modeling in the context of supervised and unsupervised few-shot learning given the Absolute-relative Learning paradigm via class, semantic and augmentation annotations.

\section{Related Work}
\label{sec:related}

Below, we describe recent one- and few-shot learning algorithms followed by semantic-based approaches.

\subsection{Learning From Few Samples}
\label{sec:related_few_shot}

For deep learning algorithms, the ability of {\em`learning from only a few examples is the desired characteristic to emulate in any brain-like system'} \cite{book_nip} is a desired operating principle which poses a challenge to typical CNNs designed for the large scale visual category recognition \cite{ILSVRC15}. 

\vspace{0.05cm}
\noindent{\textbf{One- and Few-shot Learning }} has been  studied widely  in computer vision in both shallow \cite{miller_one_example,Li9596,NIPS2004_2576,BartU05,fei2006one,lake_oneshot} and deep learning scenarios \cite{koch2015siamese,vinyals2016matching,snell2017prototypical,finn2017model,snell2017prototypical,sung2017learning,sosn}. 

Early works \cite{fei2006one,lake_oneshot} propose one-shot learning methods motivated by the observation that humans can learn new concepts from very few examples. Siamese Network \cite{koch2015siamese} presents a two-streams convolutional neural network approach which generates image descriptors and learns the similarity between them. Matching Network \cite{vinyals2016matching} introduces the concept of support set and $L$-way $Z$-shot learning protocols. It  captures the similarity between one testing and several support images, thus casting the one-shot learning problem as set-to-set learning. Prototypical Networks \cite{snell2017prototypical} learns a model that computes distances between a datapoint and prototype representations of each class. Model-Agnostic Meta-Learning (MAML) \cite{finn2017model} introduces a meta-learning model trained on a variety of different learning tasks. 
Relation Net \cite{sung2017learning}
is an efficient end-to-end network for learning the relationship between testing and support images. Conceptually, this model is similar to Matching Network \cite{vinyals2016matching}. However, Relation Net leverages an additional deep neural network to learn similarity on top of the image descriptor generating network. Second-order Similarity Network (SoSN) \cite{sosn} is similar to Relation Net \cite{sung2017learning}, which consists of the feature encoder and relation network. However, approach \cite{sung2017learning} uses first-order representations for similarity learning. In contrast, SoSN investigates second-order representations to capture co-occurrences of features. Graph Neural Networks (GNN) have also been applied to few-shot learning in many recent works \cite{garcia2017few,Kim_2019_CVPR,Gidaris_2019_CVPR} achieving promising results. Finally, noteworthy are domain adaptation and related approaches which can also  operate in the small sample regime \cite{koniusz2017domain,me_museum,zhang2018zero,zhang2018model,li2020word,li2020transferring,NEURIPS2020_8c00dee2}.


\subsection{Learning from Semantic Labels}
Semantic labels are  used in various computer vision tasks \eg, object  classification, face and emotion recognition, image retrieval, transfer learning, and especially in zero-shot learning. Metric learning  often uses semantic information \eg, approach 
 \cite{bradshaw2000semantic} proposes an image retrieval system which uses semantics of images via probabilistic modeling. Approach \cite{wang2011image} presents a novel bi-relational graph model that comprises both the data graph and semantic label graph, and connects them by an additional bipartite graph built from label assignments. Approach \cite{peng2014learning} proposes a classifier based on semantic annotations and provides the theoretical bound linking the error rate of the classifier and the number of instances required for training. 
 Approach \cite{huai2018metric}  improves metric learning via the use of semantic labels 
 with different types of semantic annotations. 
 
 {Our relative learning is somewhat related to the idea using semantic information to learn metric. However, we use similarity measures to simulate realistic relation labels in supervised and unsupervised few-shot learning.}

\subsection{Multi-task Learning}
{Multi-task learning operates on a set of multiple related tasks. Approach \cite{argyriou2007multi} treats the multi-task learning as a convex iterative problem. Approach \cite{Kendall_2018_CVPR}  considers the homoscedastic uncertainty of each task to weight multiple loss functions while HallNet \cite{hall_net} learns old-fashioned descriptors as auxiliary tasks for action recognition.

In contrast, we focus on how to refine the backbone by learning from class concepts and relations to address the high-level few-shot learning task.}

\comment{
\begin{figure*}[t]
    \centering
    \vspace{-0.3cm}
    \includegraphics[width=0.9\linewidth]{images/separate-models.pdf}%
    \vspace{-0.2cm}
    \caption{The pipeline: standard Relation Learning, Relative Learning and Absolute Learning. The original SoSN uses the binary labels for supervision while we introduce different types of labelling which can be though of as soft \vs hard, and relative \vs absolute. Relative Learning employs semantic annotations (\eg, attributes, word2vec) to capture more realistic object relations while Absolute Learning leverages an extra branch of classifier to predict object annotations \eg, class label, attributes, and word2vec embeddings.}
    \label{fig:pipe-comp}%
    \vspace{-0.7cm}
\end{figure*}}

\section{Background}
The concept of few-shot learning and the standard pipeline for few-shot learning are described next.

\subsection{Relation Learning}
Few-shot learning model typically consists of two parts: (i) feature encoder and (ii) relation module \eg, a similarity network or a classifier. 
Below we take the two-stage `feature encoder-relation network' \cite{sung2017learning,sosn} as an example to elaborate on main aspects of few-shot learning pipelines. 

A basic relation network \cite{sung2017learning,sosn} contains 2-4 convolutional blocks and 2 fully-connected layers. 
Let us define the feature encoding network as  $f\!:(\mbr{W\!\times\!H}; \mbr{|\tF|})\!\shortrightarrow\!\mbr{K\!\times\!N}$, where $W$ and $H$ denote the width and height of an input image, $K$ is the length of feature vectors (number of filters), $N\!=\!N_W\!\cdot\!N_H$ is the total number of spatial locations in the last convolutional feature map. For simplicity, we denote an image descriptor by $\mPhi\!\in\!\mbr{K\!\times\!N}$, where $\mPhi\!=\!f(\mX; \tF)$ for an image $\mX\!\in\!\mbr{W\!\times\!H}$ and $\tF$ are the parameters-to-learn of the encoding network.

The relation network is denoted by  $r\!:(\mbr{K'\!}; \mbr{|\tR|})\!\shortrightarrow\!\mbr{}$. Typically, we write $r(\vpsi; \tR)$,  where $\vpsi\!\in\!\mbr{K'}\!$, whereas $\tR$ are the parameters-to-learn of the relation network.

\subsection{Supervised Few-shot Learning}

\comment{\begin{figure*}[t]
    \centering
    \includegraphics[width=0.9\linewidth]{images/k-func.pdf}
    \caption{\small The two element-wise measurement functions investigated in our work. The plots are the function curves w.r.t different p values. Right figure is the semantic-guided class relations. These relations are derived from attributes with Gaussian kernel function of different $\sigma$ values for 100 \textit{mini}Imagenet classes. It can be seen the semantic annotation can simulate more realistic object relations than binary labels.}
    \vspace{-0.5cm}
    \label{fig:soft_label}
\end{figure*}}

For the supervised $L$-way $Z$-shot problem, we assume some support images $\{\mX_s\}_{s\in\mathcal{W}}$ from set $\mathcal{W}$ and their corresponding image descriptors $\{\mPhi_s\}_{s\in\mathcal{W}}$ which can be considered as a $Z$-shot descriptor. Moreover, we assume one query image $\mX_q\!$ with its image descriptor $\mPhi_q$. Both the $Z$-shot and the query descriptors belong to one of $L$ classes in the subset $\mathcal{C}^{\ddag}\!\equiv\!\{c_1,\cdots,c_L\}\!\subset\!\idx{C}\!\equiv\!\mathcal{C}$. The $L$-way $Z$-shot learning step can be defined as learning similarity:
\begin{equation}
\zeta_{sq}=r\left(\vartheta\!\left(\{\mPhi_s\}_{s\in\mathcal{W}},\mPhi^*_q\!\right),\tR\right),
\end{equation}
where $\zeta$ refers to similarity prediction of given support-query pair, $r$ refers to the relation network, and $\tR$ denotes network parameters that have to be learnt. $\vartheta$ is the relation operator on features of image pairs: we simply use concatenation.

Following approaches \cite{sung2017learning,sosn}, the Mean Square Error (MSE) is employed as the objective function:
\begin{align}
\vspace{-0.2cm}
&L\!\!=\!\!\sum\limits_{c\in\mathcal{C}^{\ddag}}\!\sum\limits_{c'\in\mathcal{C}^{\ddag}} \left(r\left(\!\{\mPhi_s\}_{s\in\mathcal{W}_c},\mPhi_{q\in\mathcal{Q}: \ell(q)=c'},\tR\right)\!-\!\delta\!\left(c\!-\!c'\right)\!\right)^2\!\!,\nonumber\\
&\qquad\text{ where }\; \mPhi_s\!=\!f(\mX_s; \tF) \;\text{ and }\; \mPhi_q\!=\!f(\mX_q; \tF).
\vspace{-0.2cm}
\end{align}

In the above equation, $\mathcal{W}_c$ is a randomly chosen set of support image descriptors of class $c\!\in\!\mathcal{C}^{\ddag}$, $\mathcal{Q}$ is a randomly chosen set of $L$ query image descriptors so that its consecutive elements belong to the consecutive classes in $\mathcal{C}^{\ddag}\!\equiv\!\{c_1,\cdots,c_L\}$. $\ell(q)$ corresponds to the label of $q\!\in\!\mathcal{Q}$. Lastly, $\delta$ refers to the  indicator function equal 1 if its argument is 0.

\subsection{Unsupervised Few-shot Learning}
There are no class annotations that can be directly used for relation learning in the unsupervised setting. However, the popular self-supervised contrastive learning  captures self-object relations by learning the similarity between different augmentations of the same image. Thus, we build our unsupervised few-shot learning pipeline based on contrastive learning. Given two image inputs $\mX$ and $\mY$, we  apply random augmentations on these images \eg, rotation, flip, resized crop and color adjustment via operator $\text{Aug}(\cdot)$, which samples these transformations according to a uniform distribution. We obtain a set of $M$ augmented images:
\begin{align}
    \hat{\mX}_i \sim \text{Aug}(\mX),\;\hat{\mY}_i \sim \text{Aug}(\mY),\;i\in\{1,\cdots,M\}.   
\end{align}
We pass augmented images to the feature encoder $f$ to get feature descriptors  and obtain relation predictions $\zeta,\zeta^*\!\in\mbr{M\!\times\!M}$ from relation network $r$ for augmented samples of $\mX$ and $\mY$, respectively, as well as relation predictions $\zeta'\!\in\mbr{M\!\times\!M}$ evaluated between augmented samples of $\mX$ and $\mY$:
\begin{align}
\label{eq:enc}
&\mPhi_{i} = f(\hat{\mX}_i; \tF),\; \mPhi^*_j = f(\hat{\mY}_j; \tF),\;i,j\in\{1,\cdots,M\}, \\
&\!\!\!\!\!\!\!\!\!\!\zeta_{ij}\!=\!r\left(\mPhi_i,\mPhi_j; \tR\right),\; \zeta'_{ij}\!=\! r\left(\mPhi_i,\mPhi^*_j; \tR\right),\; \zeta^*_{ij}\!=\! r\left(\mPhi^*_i,\mPhi^*_j; \tR\right).\nonumber
\end{align}
Lastly, we minimize the contrastive loss $L_{urn}$ \wrt $\tF$ and $\tR$ in order to push closer augmented samples generated from the same image ($\mX$ and $\mY$, resp.) and push away augmented samples generated from pairs images $\mX$ and $\mY$:
{
\begin{align}
     L_{urn}=\,\parallel\!\boldsymbol{\zeta}-1\!\parallel^2_F + \parallel\!\boldsymbol{\zeta}^*\!-1\!\parallel^2_F 
     +\parallel\!\boldsymbol{\zeta}'\!\parallel^2_F.\label{eq:urn}
\end{align}}

\vspace{-0.4cm}
In practice, we sample a large number of image pairs $\mX$ and $\mY$ with the goal of minimizing Eq. \eqref{eq:urn}.

\section{Approach}
Below, we firstly explain the Relative Learning and Absolute Learning modules followed by the  introduction of the Absolute-relative Learning pipeline. We note that all  auxiliary information \eg, attributes and word2vec embeddings are  used in the label space (not as extra inputs). 
%

Given images $\mathbf{X}_i$ and $\mathbf{X}_j$, we  feed them into the feature encoder $f$ to get image representations $\mPhi_i = f(\mathbf{X}_i; \tF)$ and $\mPhi_j = f(\mathbf{X}_j; \tF)$, where $\tF$ are the parameters of feature encoder. Subsequently, we perform our proposed Relative Learning and Absolute Learning on $\mPhi_i$ and $\mPhi_j$. 

\subsection{Relative Learning}
In conventional few-shot learning, binary class labels are employed to train the CNNs in order to model the relations between pairs of images. However, labeling such pairs as similar/dissimilar (\ie, $\{0,1\}$) cannot fully reflect the actual relations between objects. 


In this paper, we take a deeper look at how to represent relations in the few-shot learning scenario. To better exploit class relations in the label space, we employ semantic annotations \eg, attributes and word2vec. Based on these semantic annotations, we  investigate how semantic relation labels influence the final few-shot learning performance. 

\begin{figure*}[t]
    \centering
    \includegraphics[width=\linewidth]{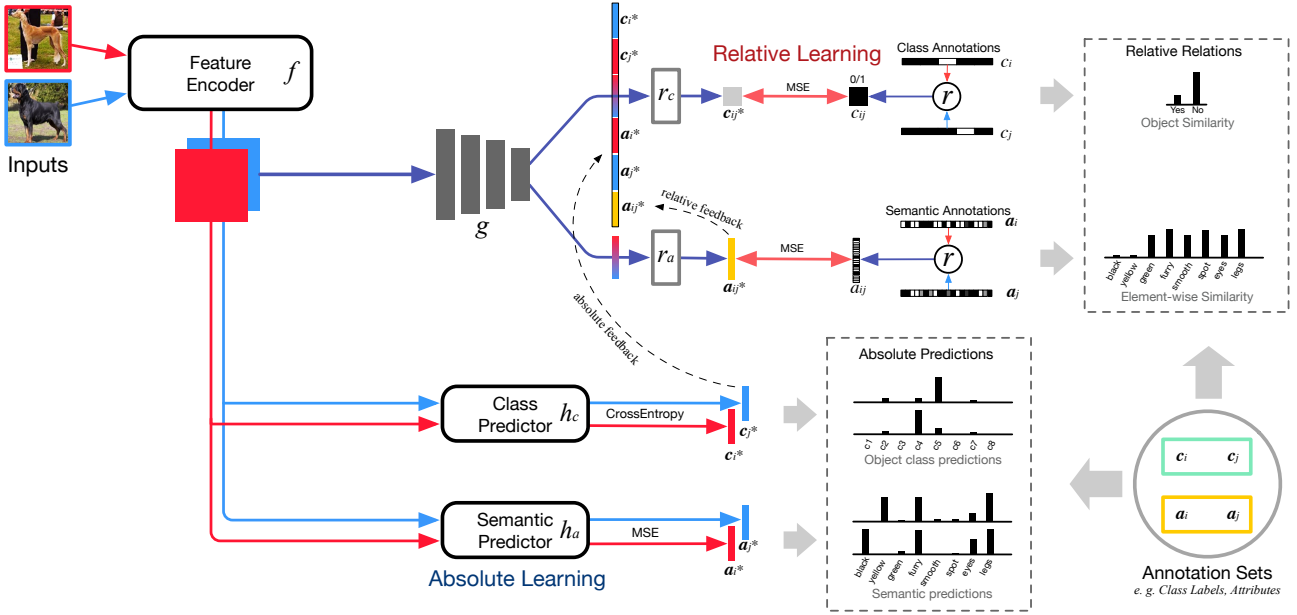}%
    \vspace{-0.3cm}
    \caption{\small The proposed pipeline for our Absolute-relative Learning (supervised setting). It consists of three blocks, namely (i) feature encoder to extract the image representations, (ii) Absolute Learning module to enhance the feature quality with auxiliary supervision, (iii) Relative Learning module to learn image relations based on multi-modal relation supervisions. With our Absolute-relative Learning, we want to both learn if the two objects share the same label and how similar they are semantically \eg, in terms of shared visual attributes.}%
    \label{fig:arl}%
    \vspace{-0.3cm}
\end{figure*}

Figure \ref{fig:arl} (bottom right corner) shows that the classic relation learning can be viewed as an intersection (or relation) operation over the original class labels. Thus, we apply intersection on the semantic annotations to obtain the relative semantic information for the relative supervision, which can contribute to obtaining more realistic image relations in the label space. Let us   denote the class labels and attributes of image $\textbf{X}_i$ as $c_i, \mathbf{a}_i$. 
Given two samples $\textbf{X}_i$ and $\textbf{X}_j$ with their class labels $c_i, c_j$ (and one-hot vectors $\mathbf{c}_i, \mathbf{c}_j$) and attributes $\mathbf{a}_i, \mathbf{a}_j$, we obtain the binary relation label $\hat{c}_{ij}$ which represents if the two images are from the same class. We also have semantic relation label $\hat{a}_{ij}$ which represents attributes shared between $\textbf{X}_i$ and $\textbf{X}_j$. 
Semantic annotations often contain continuous rather than binary values. 
Thus, we use the RBF function with the $\ell^p_p$ norm. Specifically, we obtain:
%
\begin{align}
\label{eq6}
\!\!\!\!\hat{c}_{ij}\!=\!c_i\!\wedge\!c_j\!=\!\delta(\mathbf{c}_i\!-\!\mathbf{c}_j) \text{ and } \hat{a}_{ij}\!=\!e^{-||\mathbf{a}_i-\mathbf{a}_j||_p^p},
\end{align}

If we train the network only with $\hat{c}_{ij}$, it becomes the basic few-shot learning. However, the simultaneous use of $\hat{c}_{ij}$ and $\hat{a}_{ij}$ for similarity learning should yield smoother similarity decision boundaries.

To learn from multi-modal relative supervisions, we apply a two-stage learner consisting of a shared part $g$ and respective parts $r$. Let us denote the class and semantic relative learners as $r_c$ and $r_a$. To make relative predictions, we firstly apply the relation operator $\vartheta$ over $\mPhi_i$ and $\mPhi_j$ (concatenation along the channel mode), and feed such a relation descriptor into $g$ (4 blocks of Conv-BN-ReLU-MaxPool) to obtain the refined pair-wise representation $\vpsi_{ij}$:
\begin{equation}
    \vpsi_{ij} = g(\vartheta(\mPhi_i, \mPhi_j);\tG).
\end{equation}

Subsequently, we feed  $\vpsi_{ij}$ into learners $r_c$ and $r_a$ to get class- and semantics-wise relation predictions $\hat{c}^*_{ij}$ and  $\hat{a}^*_{ij}$:
\begin{align}
    \hat{c}^*_{ij} = r_c(\vpsi_{ij}; \tR_c) \text{ and } \hat{a}^*_{ij} = r_a(\vpsi_{ij}; \tR_s),
\end{align}
where $\tR_c$ and  $\tR_s$ refer to the parameters of $r_c$ and  $r_a$, respectively. The objectives for class- and semantic-wise relative learners are:
\begin{align}
&L_{relc}=\sum\limits_i\!\sum\limits_j\! \left(r_c\left(\vpsi_{ij};\tR_c\right)-\hat{c}_{ij}\right)^2, \\
&L_{rels}=\sum\limits_i\!\sum\limits_j\! \left(r_a\left(\vpsi_{ij};\tR_s\right)-\hat{a}_{ij}\right)^2. 
\vspace{-0.3cm}
\end{align}


\subsection{Absolute Learning}
In contrast to Relative Learning which applies the relative labels to learn similarity, Absolute Learning refers to the strategy in which the network learns predefined object annotations \eg, class labels, attributes, \etc. 
The motivation behind the Absolute Learning is that current few-shot learning pipelines use the  relation labels as supervision which prevents the network from capturing objects concepts. In other words, the network knows if the two objects are similar (or not) but it does not know what these objects are.

Branches for Absolute Learning are shown in Figure \ref{fig:arl}. In this paper, we apply an additional network branch following the feature encoder to learn the absolute object annotations.
\begin{figure*}[t]
    \centering
    \includegraphics[width=\linewidth]{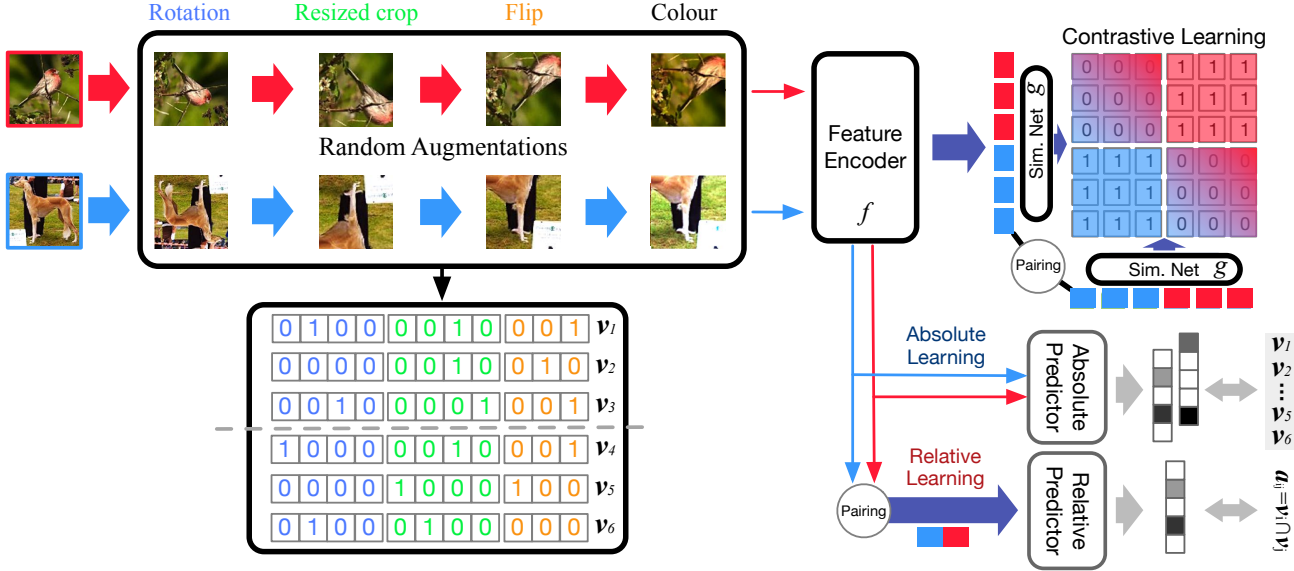}%
    \vspace{-0.3cm}
    \caption{\small The proposed pipeline for Absolute-relative Learning (unsupervised setting). In contrast to supervised ArL that uses class and semantic annotations in absolute and relative learners, we apply a random augmentation sequence to augment unlabeled datapoints, and we store the augmentation keys as instance annotations.}%
    \label{fig:uarl}
    \vspace{-0.3cm}
\end{figure*}
Firstly, consider the class prediction as an example. Once we obtain the image representation $\mPhi_i$ given image $\mathbf{X}_i$, we feed it into the class absolute learner  $h_c$ with parameter $\tH_c$.:
%
\begin{equation}
    \mathbf{c}^*_i = h_c(\mPhi_i; \tH_c).
    \vspace{-0.2cm}
\end{equation}
Subsequently, we apply the cross-entropy loss to train the class absolute learner ($l^c$ is the target class integer):
\vspace{-0.3cm}
\begin{equation}
    L_{absc} = -\sum\limits_i^N log(\frac{exp(c^*_i[l^c_{i}])}{\sum\limits_{j} exp(c^*_i[j])}).
\end{equation}

\vspace{-0.3cm}
\noindent For the semantic absolute learner, we use the MSE loss by feeding $\mPhi_i$ into $h_a$: 
\vspace{-0.3cm}
\begin{equation}
    \mathbf{a}^*_i = h_a(\mPhi_i; \tH_s),
\end{equation}
\begin{equation}
    L_{abss} = \frac{1}{N} \sum\limits_i^N || \mathbf{a}_i - \mathbf{a}^*_i ||_2^2.
\end{equation}

The Absolute Learning module may appear somewhat similar to self-supervised learning applied to  few-shot learning. 
However, we use discriminators to classify different types of object annotations while the typical self-supervision recognises the patterns of image transformations. We believe our strategy helps refine the feature encoder to capture both the notion of similarity as well as concrete object concepts.

\subsection{Absolute-relative Learning}
For our Absolute-relative Learning (ArL), we simultaneously train the relation network with relative object similarity labels, and introduce an auxiliary task which learns specific object labels. The pipeline of ArL is shown in Figure \ref{fig:arl} which highlights that the ArL model uses the auxiliary semantic soft labels to train the relation network to capture more realistic image relations while employing auxiliary predictor branches to infer different types of object information, thus refining  the feature representations and the feature encoder. 

In addition to merging the absolute/relative learners, we  introduce several connections from the outputs of absolute and relative learners wired to relative learners to promote the original relation learner, which does not require absolute labels or semantic labels at the testing time. In contrast,  multi-modal learning  needs all modalities in the testing step.

Figure \ref{fig:uarl} shows the Absolute-relative Learning pipeline (unsupervised setting). As the supervised ArL, the unsupervised ArL pipeline consists of absolute and relative learners. However, the annotations used during the training phase are self-supervised augmentation keys, not 
class labels.

Let $l$ denote the number of layers in $g$. We apply $\vartheta$ over the intermediate descriptor $\vpsi_{ij}^{(l-1)}\!$, which is the ($l$-1)-th layer of $g$, and absolute predictions $\mathbf{c}^*_i, \mathbf{c}^*_j, \mathbf{a}^*_i, \mathbf{a}^*_j$. We call this operation the absolute feedback:
\begin{equation}
\!\!\!\!\!\!\!\!\!\!\!\hat{\vpsi}_{ij}^{(l-1)}\!=\! \vartheta\left(\vpsi_{ij}^{(l-1)},\mathbf{c}^*_i),\mathbf{c}^*_j),\mathbf{a}^*_i,\mathbf{a}^*_j\right).
\end{equation}
%
%
We use   $\hat{\vpsi}_{ij}$ from the last layer of $g$ to train the semantic relative learner $r_a$:
\begin{align}
\!&\hat{\vpsi}_{ij} = \hat{\vpsi}_{ij}^{(l)} = g^{(l)}\big(\hat{\vpsi}_{ij}^{(l-1)}\big),\\
\!&L_{rels}=\sum\limits_i\!\sum\limits_j\! \left(\!r_a\big(\hat{\vpsi}_{ij};\tR_s\big)-\hat{a}_{ij}\right)^2\!\!,\;\hat{a}_{ij}\!=\!e^{-||\mathbf{a}_i-\mathbf{a}_j|_p^p}.\nonumber
\end{align}
%
\begin{table*}[t]
\centering
\caption{Evaluations on the \textit{mini}Imagenet dataset (5-way acc. given) for the ArL in supervised and unsupervised settings. (`U-' refers to the unsupervised FSL.)
}
\vspace{-0.2cm}
\label{table1}
\makebox[\textwidth]{
\setlength{\tabcolsep}{1em}
\renewcommand{\arraystretch}{1}
\fontsize{9}{10}\selectfont
\begin{tabular}{lccc}
\toprule
Model & Backbone & 1-shot & 5-shot \\ \midrule
& \multicolumn{3}{c}{Supervised Few-shot Learning} \\ \midrule
\textit{Matching Nets} \cite{vinyals2016matching} & - & $43.56 \pm 0.84 $ & $55.31 \pm 0.73 $  \\
\textit{Meta Nets} \cite{munkhdalai2017meta} & - & $49.21 \pm 0.96 $ & - \\
\textit{PN} \cite{snell2017prototypical} & Conv-4-64 & $49.42 \pm 0.78 $ & $68.20 \pm 0.66 $ \\
\textit{MAML} \cite{finn2017model} & Conv-4-64 & $48.70 \pm 1.84 $ & $63.11 \pm 0.92 $ \\
\textit{RN} \cite{sung2017learning} & Conv-4-64 & $51.36 \pm 0.82 $ & $66.12 \pm 0.70 $  \\
\textit{SoSN} \cite{sosn} & Conv-4-64 & $53.73\pm 0.83 $ & $68.58 \pm 0.70$ \\
\textit{SoSN} \cite{sosn} & ResNet-12 & $59.01\pm 0.83 $ & $75.49 \pm 0.68$ \\
\textit{MAML++} \cite{antoniou2018train} & Conv-4-64 & $52.15 \pm 0.26 $ & $68.32 \pm 0.44$ \\
\textit{MetaOptNet} \cite{lee2019meta} & ResNet-12 & $62.64\pm 0.61 $ & $78.63 \pm 0.46$\\
\arrayrulecolor{black} \cdashline{1-4}[1pt/3pt]
\textit{PN + ArL} & Conv-4-64 & $53.93 \pm 0.65$& $69.68 \pm 0.45$ \\
\textit{RN + ArL} & Conv-4-64 & $53.79 \pm 0.68$& $68.86 \pm 0.43$ \\
\textit{SoSN + ArL} & Conv-4-64 & $57.48 \pm 0.65$ & $72.64 \pm 0.45$ \\
\textit{SoSN + ArL} & ResNet-12 & ${61.36 \pm 0.67}$ & ${78.95 \pm 0.42}$ \\
\textit{MetaOptNet + ArL} & ResNet-12 & ${65.21\pm 0.58}$ & ${80.41 \pm 0.49}$  \\
\hline
& \multicolumn{3}{c}{Unsupervised Few-shot Learning} \\
\hline
\textit{Pixel (Cosine)} & - & $23.00$ & $26.60$  \\
\textit{BiGAN ($k_{nn}$)} \cite{bigan} & - & $25.56$ & $31.10$  \\
\textit{BiGAN (cluster matching)} \cite{bigan} & - & $24.63$ & $29.49$  \\
\textit{DeepCluster ($k_{nn}$)} \cite{deepcluster} & - & $28.90$ & $42.25$  \\
\textit{DeepCluster (cluster matching)} \cite{deepcluster} & - & $22.20$ & $23.50$  \\
\textit{UMTRA } \cite{umtra} & Conv-4-64 & $39.91$ & $ 50.70$  \\
\textit{CACTUs} \cite{cactu} & Conv-4-64 & $39.94$ & $ 54.01$  \\
\arrayrulecolor{black} \cdashline{1-4}[1pt/3pt]
\textit{U-RN} & Conv-4-64 & $35.14 \pm 0.91$ & $44.10 \pm 0.88$ \\
\textit{U-PN} & Conv-4-64 & $35.85 \pm 0.85$ & $48.01 \pm 0.82$ \\
\textit{U-SoSN} & Conv-4-64 & $37.94 \pm 0.87$ & $50.95 \pm 0.81$ \\
\textit{U-RN + ArL} & Conv-4-64 & $36.37 \pm 0.92$ & $46.97 \pm 0.86$ \\
\textit{U-PN + ArL} & Conv-4-64 & $38.76 \pm 0.84$ & $51.08 \pm 0.84$ \\
\textit{U-SoSN + ArL} & Conv-4-64 & ${41.13 \pm 0.84}$ & ${55.39 \pm 0.79}$ \\
\textit{U-SoSN + ArL} & ResNet-12 & $41.08 \pm 0.83$ & ${57.01 \pm 0.79}$ \\
\bottomrule
\end{tabular}}
\vspace{-0.4cm}
\end{table*}
\begin{table}[b]
\centering
\vspace{-0.3cm}
\caption{Evaluations on the CUB-200-2011 and Flower102. (5-way acc. given).}
\vspace{-0.2cm}
\label{table2}
\makebox[\linewidth]{
\setlength{\tabcolsep}{1em}
\renewcommand{\arraystretch}{1}
\fontsize{8.5}{9.5}\selectfont
\begin{tabular}{lcccc}
\toprule
& \multicolumn{2}{c}{CUB-200-2011} & \multicolumn{2}{c}{Flower102} \\
Model & 1-shot & 5-shot & 1-shot & 5-shot \\ \midrule
& \multicolumn{4}{c}{Supervised Few-shot Learning} \\
\hline
\textit{PN}  \cite{snell2017prototypical} & $37.42 $ & $51.57$ & $62.81$ & $82.11$  \\
\textit{RN} \cite{sung2017learning} & $40.56 $ & $53.91$ & $68.26$ & $80.94$  \\
\textit{SoSN}  \cite{sosn}& $46.72$ & $60.34$ & $71.90$ & $84.87$ \\
\textit{RN + ArL} & ${44.53}$ & ${58.76}$  & ${71.12}$  & ${83.49}$  \\
\textit{SoSN - RL(cls.)}  \cite{sosn}& $46.72$ & $60.34$ & $71.90$ & $84.87$ \\
\textit{SoSN - RL(att.)} & ${49.24}$  & ${64.04}$  & ${74.96}$ & ${87.21}$\\
\textit{SoSN - AL(cls.)} & $46.88$ & $60.90$ & $72.97$ & $85.35$ \\
\textit{SoSN - AL(att.)} & ${48.85}$  & ${63.64}$ & ${74.31}$ & $86.97$  \\
\textit{SoSN + ArL} & $\mathbf{50.62}$ & ${65.87}$  & ${76.21}$  & ${88.36}$ \\
\hline
& \multicolumn{4}{c}{Unsupervised Few-shot Learning} \\
\hline
\textit{BiGAN($k_{nn}$)}\cite{bigan} & $28.02$ & $30.17$ & $44.68$ & $59.12$  \\ 
\textit{U-RN} & ${29.36}$ & ${36.36}$ & ${55.54}$ & ${68.86}$ \\
\textit{U-PN} & ${29.87}$ & ${37.13}$ & ${55.36}$ & ${68.49}$ \\
\textit{U-SoSN} & ${36.89}$ & ${45.81}$ & ${61.26}$ & ${75.98}$ \\
\textit{U-RN + ArL} & ${31.27}$ & ${38.41}$ & ${57.19}$ & ${70.23}$ \\
\textit{U-PN + ArL} & ${31.58}$ & ${39.95}$ & ${57.61}$ & ${70.31}$ \\
\textit{U-SoSN + ArL} & ${37.93}$ & ${51.55}$ & ${69.14}$ & ${84.10}$ \\
\bottomrule
\end{tabular}}
\end{table}

\vspace{-0.3cm}
Let $\hat{\mathbf{a}}^*_{ij}$  denote the outputs of semantic relative learner. Then we apply the relative feedback by combining $\hat{\vpsi}_{ij}$ and $\hat{\mathbf{a}}^*_{ij}$ to promote the training of class relative learner $r_c$:
%
\begin{align}
  & \hat{\vpsi}^*_{ij} = \vartheta(\hat{\vpsi}_{ij},\hat{\mathbf{a}}^*_{ij}),\\
 &   L_{relc}=\sum\limits_i\!\sum\limits_j\! \left(\!r_c\big(\hat{\vpsi}^*_{ij};\tR_c\big)-\hat{c}_{ij}\right)^2\!\!,\; \hat{c}_{ij}\!=\!\delta(\mathbf{c}_i\!-\!\mathbf{c}_j).\nonumber
\end{align}
\vspace{-0.3cm}

\noindent We minimize the following objective for ArL:
\begin{equation}
    \min \quad L_{relc} + \alpha L_{rels} + \beta L_{absc} + \gamma L_{abss}.
\end{equation}
where $(\alpha, \beta, \gamma)\!\in\![0.001; 1]^3$ are hyper-parameters that control the impact of each learner and are estimated with 20 steps of the HyperOpt package \cite{bergstra2015hyperopt}  on a given validation set. Nullifying $\alpha$, $\beta$ or $\gamma$ disables corresponding losses. 

\section{Experiments}
Below, we demonstrate the usefulness of our approach by evaluating it on the \textit{mini}Imagenet \cite{vinyals2016matching}, fine-grained CUB-200-2011 \cite{WahCUB_200_2011} and Flower102 \cite{Nilsback08} datasets. 
Figure \ref{fig:arl} presents  our ArL with the two-stage relation learning pipeline but ArL applies to any few-shot learning models with any type of base learners (\eg, nearest neighbour discrimination, relation module, multi-class linear classifier, \etc). The core objective of ArL is to improve the representation quality. Thus, we employ the classic baseline models, \ie, Prototypical Net (PN) \cite{snell2017prototypical}, Relation Net \cite{sung2017learning}, SoSN \cite{sosn}, MetaOptNet \cite{lee2019meta}, \etc, as our baseline models to evaluate our Relative Learning, Absolute Learning  and the Absolute-relative Learning  in both supervised and unsupervised settings. The Adam solver is used for model training. We set the initial learning rate to be 0.001 and decay it by 0.5 every 50000 iterations. %
We evaluate ArL 
on RelationNet (RN) \cite{sung2017learning}, Prototypical Net (PN) \cite{snell2017prototypical}, Second-order Similarity Network (SoSN) \cite{sosn} and MetaOptNet \cite{lee2019meta}. 
For augmentations in the unsupervised setting, we randomly apply resized crop (scale 0.6--1.0, ratio 0.75--1.33), horizontal+vertical flips, rotations ($0$--$360^{\circ}$), and color jitter.

\subsection{Datasets}
Below, we describe our  setup, standard  and fine-grained datasets with semantic annotations and  evaluation protocols.

\vspace{0.05cm}
\noindent\textbf{\textit{mini}Imagenet} \cite{vinyals2016matching} consists of 60000 RGB images from 100 classes, each class containing 600 samples. 
We follow the standard protocol \cite{vinyals2016matching} and use 80/20 classes for training/testing, and  images of size $84\!\times\!84$ for fair comparisons with other methods. For semantic annotations, we manually annotate 31 attributes for each class. We also leverage word2vec extracted from GloVe as the class embedding.

\vspace{0.05cm}
\noindent\textbf{Caltech-UCSD-Birds 200-2011 (CUB-200-2011)} \cite{WahCUB_200_2011} has 11788 images of 200 bird species. 100/50/50 classes are randomly selected for meta-training,  meta-validation and  meta-testing. 312 attributes are provided for each class.

\vspace{0.05cm}
\noindent\textbf{Flower102} \cite{Nilsback08} is a fine-grained category recognition dataset that contains 102 classes of various flowers. Each class consists of 40-258 images. We randomly select 60 meta-train classes, 20 meta-validation classes and 22 meta-test classes. 1024 attributes are provided for each class.

\begin{table}[b]
    \centering
    \vspace{-0.3cm}
    \caption{Ablation study of the impact of  different annotations (\eg,   class labels, attributes) on ArL.}
    \label{table_ablation}
    \makebox[\linewidth]{
    \setlength{\tabcolsep}{0.9em}
    \renewcommand{\arraystretch}{1}
    \fontsize{8.5}{8}\selectfont
    \begin{tabular}{ccccccc}
        \toprule
        & \multicolumn{2}{c}{Rel. Learn.} & \multicolumn{2}{c}{Abs. Learn.} & \multicolumn{2}{c}{Top-1 Acc.} \\
        Baseline & cls. & att. & cls. & att. & 1-shot & 5-shot \\ \hline
        \multirow{4}{*}{RN} & \checkmark & & & & 51.36 & 65.32 \\
        & & \checkmark & & & 52.38 & 66.74 \\
        & & & \checkmark & & 51.41 & 66.01 \\
        & & & & \checkmark & 52.35 & 66.53 \\\hline
        \multirow{4}{*}{SoSN} & \checkmark & & & & 53.73 & 68.58 \\
        & & \checkmark & & & 55.56 & 70.97 \\
        & & & \checkmark & & 55.12 & 70.91 \\
        & & & & \checkmark & 55.31 & 71.03 \\
        \bottomrule
    \end{tabular}}
\end{table}

\subsection{Performance Analysis}

\begin{figure}[t]
    \vspace{-0.2cm}
    \centering
    \includegraphics[width=\linewidth]{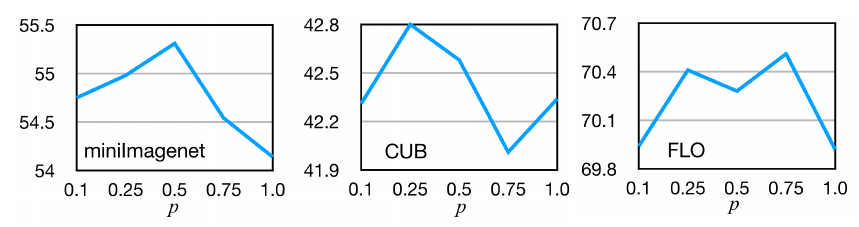}
    \vspace{-0.7cm}
    \caption{The validation of $p$ given the semantic similarity measure function $e^{-||\mathbf{a}_i-\mathbf{a}'_i||_p^p}$ on selected datasets.}
    \label{fig:fig4}
\end{figure}

\begin{figure}[t]
    \vspace{-0.3cm}
    \centering
    \includegraphics[width=\linewidth]{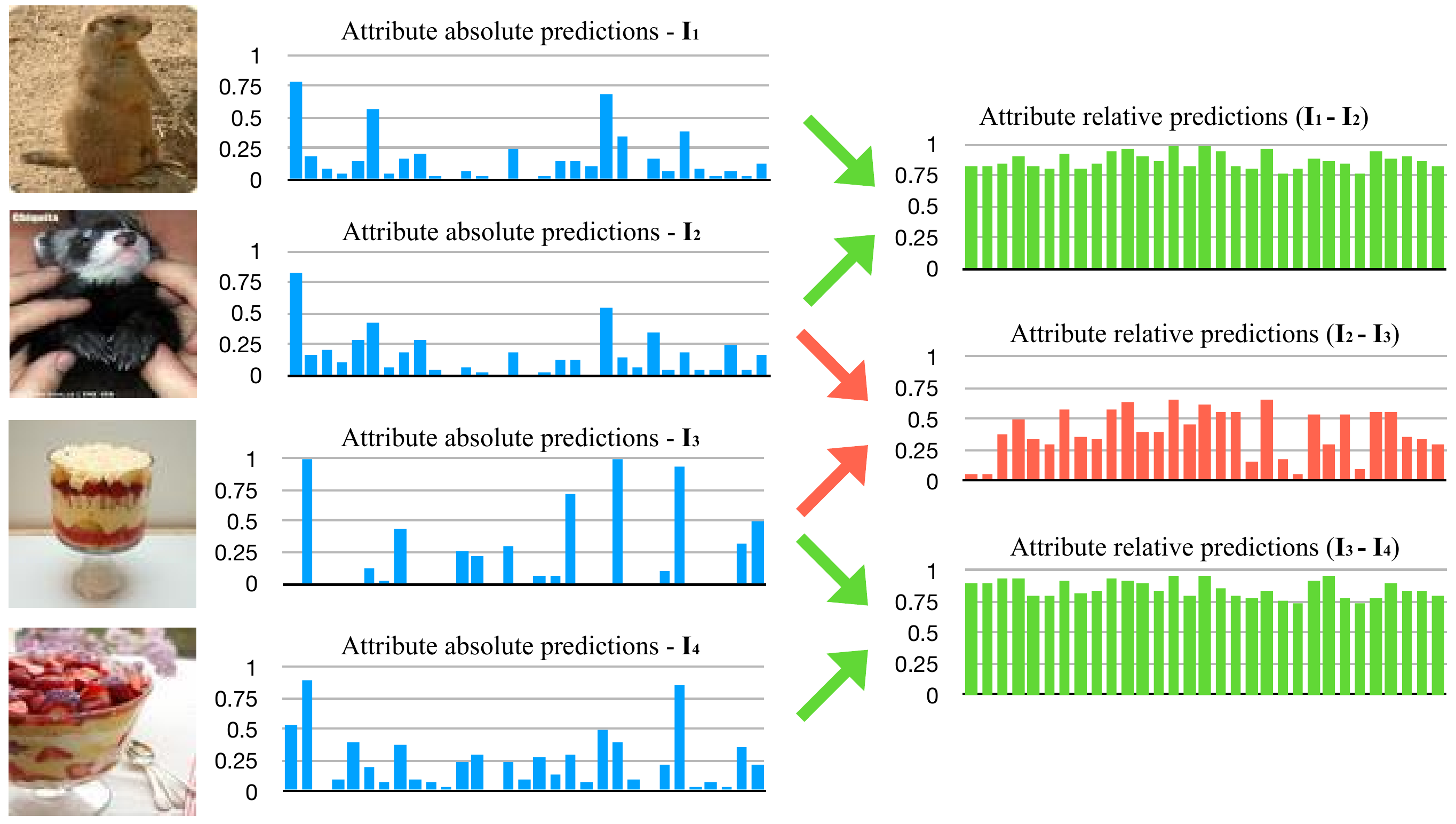}
    \vspace{-0.7cm}
    \caption{Visualization of semantic absolute and relative predictions which shows how their bins relate.}
    \vspace{-0.3cm}
    \label{fig:fig5}
\end{figure}

\noindent\textbf{Absolute-relative Learning (ArL).} 
Table \ref{table1} shows that Absolute-relative Learning (ArL) effectively improves the performance on all datasets. On \textit{mini}Imagenet, SoSN+ArL improve the 1- and 5-shot performance by $3.6\%$ and $4.1\%$, MetaOptNet+ArL improves the performance by 2.6\% and 1.8\% respectively. Not in the table,  DeepEMD \cite{Zhang_2020_CVPR} (ResNet-12) and LaplacianFSL \cite{ziko2020laplacian} (ResNet-18) scored 65.91\% and 66.41\% (1-shot prot.) In contrast, DeepEMD+ArL and LaplacianFSL+ArL scored 67.24\% and 68.07\%. 

For fine-grained datasets, CUB-200-2011 and Flower102 in Table \ref{table2}, SoSN+ArL improves the 1- and 5-shot accuracy by $1.4\%$ and $1.6\%$. 
For unsupervised learning, ArL with SoSN brings 3.5\% and 4.4\% gain on \textit{mini}Imagenet, 1.0\% and 5.7\% gain on CUB-200-2011, 7.9\% and 8.1\% gain on Flower102 for 1- and 5-shot learning, respectively.  Our unsupervised U-SoSN+ArL often outperforms recent supervised methods on fine-grained classification datasets. 

\vspace{0.05cm}
\noindent\textbf{Visualization.} Below, we visualize absolute and relative semantic predictions to explain how such an  information can be used. 
As shown in Fig. \ref{fig:fig5}, we randomly select 4 images from two classes, among which $I_1$ and $I_2$ belong to one class, and $I_3$ and  $I_4$ belong to another class. Figure \ref{fig:fig5} shows that the semantic absolute predictions of images from the same class have more consistent distributions, the relative predictions over same-class image pairs have high responses to the same subset of bins. Predictions over  images from disjoint classes result in smaller intersection of corresponding peaks.

\vspace{0.05cm}
\noindent\textbf{Ablations on absolute and relative learners.} 
Figure \ref{fig:fig4} shows  results \wrt $p$ from Eq. \ref{eq6}. 
Table \ref{table_ablation} shows how different absolute and relative learners affect  few-shot learning results on \textit{mini}Imagenet. For example, for the SoSN baseline, the attribute-based absolute and relative learners work the best among all absolute and relative learning modules. 

\vspace{0.05cm}
\noindent\textbf{Relative Learning (RL).} 
Table \ref{table_ablation} (\textit{mini}Imagenet) illustrates the performance enhanced by the semantic-based relation on Relation Net, SoSN and SalNet. Results in the table indicate that the performance of few-shot similarity learning can be improved by employing the semantic relation labels at the training stage. For instance, SoSN with attribute soft label ({\em att.}) achieves $0.6\%$ and $1.7\%$ gain for 1- and 5-shot protocols, compared with the baseline ({\em SoSN}) in Table \ref{table1}. The results on CUB-200-2011 and Flower102 from Table \ref{table2} indicate similar gains. 

\vspace{0.05cm}
\noindent\textbf{Absolute Learning (AL).} Table \ref{table_ablation} shows that different absolute learning modules help improve the performance on \textit{mini}Imagenet. SoSN with the attribute predictor ({\em SoSN-AL}) achieves the best performance of $55.61$\% on 1-shot and $71.03$\%  (5-shot). 
Table \ref{table_ablation} shows that applying multiple absolute learning modules does not always further improve the accuracy. The attribute-based predictor ({\em att.}) also works the  best among all variants on CUB-200-2011 and Flower102. For instance, SoSN with the attribute-based predictor achieves $2.1\%$ and $3.3\%$ improvements on CUB-200-2011, and $2.4\%$ and $2.1\%$ improvement on Flower102 for 1- and 5-shot protocols, respectively. We note that the class predictor ({\em cls.}) does not work well on the  fine-grained classification datasets.

\begin{figure}[t]
    \vspace{-0.2cm}
    \centering
    \includegraphics[width=\linewidth]{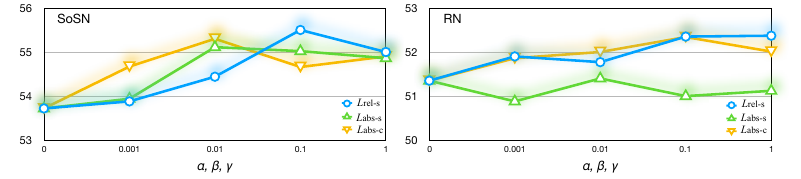}
    \vspace{-0.7cm}
    \caption{Ablations on $\alpha, \beta, \gamma$ for SoSN \cite{sosn} and RN \cite{sung2017learning} in the supervised setting. These evaluations are just an illustration as we tune parameters on the validation splits via the HyperOpt package.}
    \vspace{-0.1cm}
    \label{fig:plot}
\end{figure}

\section{Conclusions}
In this paper, we have demonstrated that  binary labels commonly used in few-shot learning  cannot capture complex class relations well, leading to inferior results. Thus, we have introduced semantic annotations to aid the modeling of more realistic class relations during network training. Moreover, we have proposed a novel 
Absolute-relative Learning (ArL) paradigm which combines the similarity learning with the concept learning, and we  extend ArL to unsupervised FSL. This surprisingly simple strategy appears to work well on all datasets in both supervised and unsupervised settings, and it perhaps resembles a bit more closely the human learning processes. In contrast to multi-modal learning, we only use semantic annotations as labels in training, and do not use them during testing. Our proposed approach achieves the state-of-the-art performance on all few-shot learning protocols. 

\vspace{0.1cm}
\noindent\textbf{Acknowledgements.} This work is in part supported by the Equipment Research and Development Fund (no. ZXD2020C2316), NSF Youth Science Fund (no. 62002371), the ANU VC's Travel Grant and CECS  Dean's Travel Grant (H. Zhang's stay at the University of Oxford).

\begin{appendices}
\section{Additional results in unsupervised setting.}
Below we supplement additional results in the unsupervised setting on two popular datasets, \textit{tiered--}Imagenet and OpenMIC.

\vspace{0.05cm}
\noindent\textbf{\textit{tiered--}Imagenet} consists of 608 classes from ImageNet. We follow the protocol that uses 351 base classes, 96 validation classes and 160 novel test classes.

\vspace{0.05cm}
\noindent\textbf{Open MIC} is the Open Museum Identification Challenge (Open MIC) \cite{me_museum}, a recent dataset with photos of various museum exhibits, \eg~paintings, timepieces, sculptures, glassware, relics, science exhibits, natural history pieces, ceramics, pottery, tools and indigenous crafts, captured from 10 museum spaces according to which this dataset is divided into 10 subproblems. In total, it has 866 diverse classes and 1--20 images per class. Following the setup in SoSN, we combine ({\em shn+hon+clv}), ({\em clk+gls+scl}), ({\em sci+nat}) and ({\em shx+rlc}) into subproblems {\em p1}, $\!\cdots$, {\em p4}, and form 12 possible pairs in which subproblem $x$ is used for training and $y$ for testing (x$\rightarrow$y).

{
\begin{table}[t]
\centering
\caption{\small Top-1 accuracy on the novel test classes of the \textit{tiered--}Imagenet dataset (5-way acc. given). Note that  `U-' variants do not use class labels during learning at all.}
\label{table_tiered}
\makebox[\linewidth]{
\fontsize{8}{9}\selectfont
\begin{tabular}{lcc}
\toprule
Model & 1-shot & 5-shot \\ \hline
\textit{$\text{MAML}$} & $51.67 \pm 1.81$ & $70.30 \pm 0.08$ \\
\textit{$\text{Prototypical Net}$} & $53.31 \pm 0.89$ & $72.69 \pm 0.74$ \\
\textit{$\text{Relation Net}$} & $54.48 \pm 0.93$ & $71.32 \pm 0.78$ \\
\textit{$\text{SoSN}$} & $58.62 \pm 0.92$ & $75.19 \pm 0.79$ \\
\hline
\textit{\text{Pixel (Cosine)}} & $27.13 \pm 0.94$ & $32.35 \pm 0.76$ \\
\textit{\text{BiGAN($k_{nn}$)}} & $29.65 \pm 0.92$ & $34.08 \pm 0.75$ \\
\textit{$\text{U-RN}$} & $37.23 \pm 0.94$ & $49.54 \pm 0.83$ \\
\textit{$\text{U-PN}$} & $38.83 \pm 0.92$ & $50.64 \pm 0.81$ \\
\textit{$\text{U-SoSN}$} & ${42.07 \pm 0.92}$ & ${56.21 \pm 0.76}$ \\
\textit{$\text{U-SoSN} + ArL$} & $\mathbf{43.68 \pm 0.91}$ & $\mathbf{58.56 \pm 0.74}$ \\

\bottomrule
\end{tabular}}
\end{table}
}

\begin{table}[t]
\centering
\caption{Ablation studies re. the impact of absolute and relative learning modules given \textit{mini}Imagenet dataset (5-way acc. with Conv-4 backbone given). We denote the same/different class relation as ({\em bin.}), attribute-based labels (relative and absolute) as ({\em att.}), {\em word2wec} embedding (relative and absolute) as ({\em w2v.}) and absolute class labeling as ({\em cls.}) RL and AL are Absolute and Relative Learners.}
\vspace{-0.3cm}
\label{table_ablation}
\makebox[\linewidth]{
\setlength{\tabcolsep}{1em}
\renewcommand{\arraystretch}{1}
\fontsize{9}{9}\selectfont
\begin{tabular}{lcc}
\toprule
Model & 1-shot & 5-shot \\ \hline
& \multicolumn{2}{c}{Relative Learning} \\
\hline
\textit{RelationNet-RL(w2v.)} & $\mathbf{53.20}$ & $66.21$  \\
\textit{RelationNet-RL(att.)} & $52.38$ & $\mathbf{66.74}$  \\
\textit{RelationNet-RL(bin. + att. + w2v.)} & $52.38$ & $66.73$  \\
\arrayrulecolor{black} \cdashline{1-3}[1pt/3pt]
\textit{SoSN-RL(w2v.)} & $54.31$ & $69.64$ \\ 
\textit{SoSN-RL(att.)} & $54.49$ & $70.21$ \\
\textit{SoSN-RL(bin. + att. + w2v.)} & $\mathbf{55.49}$ & $\mathbf{70.86}$ \\
\arrayrulecolor{black} \cdashline{1-3}[1pt/3pt]
\textit{SalNet-RL(w2v.)} & $58.15$ & $72.45$ \\ 
\textit{SalNet-RL(att.)} & $58.43$ & $72.91$ \\
\textit{SalNet-RL(bin. + att. + w2v.)} & $\mathbf{58.67}$ & $\mathbf{73.01}$ \\
\hline
& \multicolumn{2}{c}{Absolute Learning} \\
\hline
\textit{Relation Net-AL(cls.)} & $51.41$ & $66.01$\\
\textit{Relation Net-AL(att.)} & $52.35$ & $66.53$\\
\textit{Relation Net-AL(w2v.)} & $\mathbf{52.67}$ & $\mathbf{66.91}$\\
\textit{Relation Net-AL(cls.+att.+w2v.)} & $52.30 $ & $66.51 $ \\
\arrayrulecolor{black} \cdashline{1-3}[1pt/3pt]
\textit{SoSN-AL(cls.)} & $55.12$ & $70.91$\\
\textit{SoSN-AL(att.)} & $\mathbf{55.61}$ & $\mathbf{71.03}$\\
\textit{SoSN-AL(w2v.)} & $54.78$ & $70.85$\\
\textit{SoSN-AL(cls.+att.+w2v.)} & $55.40 $ & $71.02 $ \\
\arrayrulecolor{black} \cdashline{1-3}[1pt/3pt]
\textit{Salnet-AL(cls.)} & $57.98$ & $72.56$\\
\textit{SalNet-AL(att.)} & $\mathbf{58.94}$ & $\mathbf{73.12}$\\
\textit{SalNet-AL(w2v.)} & $58.36$ & $72.96$\\
\textit{SalNet-AL(cls.+att.+w2v.)} & $58.41 $ & $73.05 $ \\
\bottomrule
\end{tabular}}
\end{table}

\begin{table*}[!h]
\centering
\caption{Evaluations on the Open MIC dataset (Protocol I) (given 5-way 1-shot learning accuracies). Note that the {` U-'} variants do not use class labels during learning at all.}
\label{table_openmic}
\makebox[\linewidth]{
\renewcommand{\arraystretch}{1}
\fontsize{8}{9}\selectfont
\begin{tabular}{lcccccccccccc}
\toprule
Model & $p1\!\!\rightarrow\!p2$ & $p1\!\!\rightarrow\!p3$& $p1\!\!\rightarrow\!p4$& $p2\!\!\rightarrow\!p1$& $p2\!\!\rightarrow\!p3$ &$p2\!\!\rightarrow\!p4$& $p3\!\!\rightarrow\!p1$& $p3\!\!\rightarrow\!p2$& $p3\!\!\rightarrow\!p4$& $p4\!\!\rightarrow\!p1$& $p4\!\!\rightarrow\!p2$& $p4\!\!\rightarrow\!p3$\\
\hline
\textit{Relation Net}  & $71.1$ & $53.6$ & $63.5$ & $47.2$ & $50.6$ & $68.5$ & $48.5$ & $49.7$ & $68.4$ & $45.5$ & $70.3$ & $50.8$\\
\textit{SoSN} &  $81.4$ & ${65.2}$ & ${75.1}$ & ${60.3}$ & ${62.1}$ & ${77.7}$ & ${61.5}$ & ${82.0}$ & ${78.0}$ & ${59.0}$ & ${80.8}$ & ${62.5}$\\
\arrayrulecolor{black} \cdashline{1-13}[1pt/3pt]
\it Pixle (Cosine) &  ${56.8}$ & ${40.4}$ & ${57.5}$ & ${33.3}$ & ${35.1}$ & ${46.1}$ & ${32.3}$ & ${44.6}$ & ${45.9}$ & ${33.5}$ & ${50.1}$ & ${34.6}$\\
\it \text{BiGAN($k_{nn}$)} &  ${59.9}$ & ${43.2}$ & ${60.3}$ & ${37.1}$ & ${38.6}$ & ${50.2}$ & ${37.6}$ & ${48.2}$ & ${47.5}$ & ${38.1}$ & ${55.0}$ & ${37.8}$\\
\it U-RN &  ${70.3}$ & ${50.3}$ & ${64.1}$ & ${42.9}$ & ${48.2}$ & ${61.1}$ & ${53.2}$ & ${59.1}$ & ${55.7}$ & ${48.5}$ & ${68.3}$ & $45.2$\\
\it U-PN &  ${70.1}$ & ${49.7}$ & ${64.4}$ & ${43.3}$ & ${47.9}$ & ${60.8}$ & ${52.8}$ & ${59.4}$ & ${56.2}$ & ${49.1}$ & ${68.8}$ & $44.9$\\
\it U-SoSN &  ${78.6}$ & ${58.8}$ & ${74.3}$ & ${61.1}$ & ${57.9}$ & ${72.4}$ & ${62.3}$ & ${75.6}$ & ${73.7}$ & ${58.5}$ & ${76.5}$ & $54.6$\\
\it U-SoSN + ArL &  $\textbf{80.2}$ & $\textbf{59.7}$ & $\textbf{76.1}$ & $\textbf{62.8}$ & $\textbf{59.6}$ & $\textbf{74.4}$ & $\textbf{64.2}$ & $\textbf{78.4}$ & $\textbf{75.2}$ & $\textbf{60.1}$ & $\textbf{79.2}$ & $\textbf{57.3}$\\
\bottomrule
\end{tabular}}
\end{table*}

\vspace{0.05cm}
\noindent\textbf{Results on tiered-Imagenet.} Table \ref{table_tiered} shows that our proposed unsupervised few-shot learning strategy achieves strong results of 42.31\% and 57.21\% accuracy for 1- and 5-shot learning protocols. Though it does not outperform the recent supervised works, the performance of many prior works is not provided for this recent dataset. In general, we believe that our ArL approach boosts unsupervised learning and our unsupervised learning yields reasonable accuracy given no training labels being used in this process at all.

\vspace{0.05cm}
\noindent\textbf{Results on Open MIC.} This dataset has very limited (3-15) images for both base and novel classes, which highlights its difference to \textit{mini}Imagenet and \textit{tiered-}Imagenet whose base classes consist of hundreds of images. Table \ref{table_openmic} shows that our unsupervised variant of Second-order Similarity Network,  U-SoSN with $224\times224$ res. images outperforms the supervised SoSN on all evaluation protocols. Even without high-resolution training images, our U-SoSN outperforms the supervised SoSN on many data splits. This observation demonstrates that our unsupervised relation learning is beneficial and practical in case of very limited numbers of training images where the few-shot learning task is closer to the retrieval setting (in Open MIC, images of each exhibit constitute on one class). Most importantly, combining ArL with unsupervised SoSN boosts results further by up to 4\%.

{
\section{Ablation study on absolute and relative learners.} 
Table \ref{table_ablation} (\textit{mini}Imagenet as example) illustrates that the semantic relation learner enhanced performance on Relation Net\cite{sung2017learning}, SoSN\cite{sosn} and SalNet\cite{salnet}. The results in the table indicate that the performance of few-shot similarity learning can be improved by employing the semantic relation labels at the training stage. For instance, SoSN with attribute soft label ({\em att.}) achieves $0.6\%$ and $1.7\%$ improvements for 1- and 5-shot compared to the baseline ({\em SoSN}). Table \ref{table_ablation} also demonstrates the ablation studies for absolute learning. It can be seen from the table that the attribute predictor works the best among all options except for SoSN, and applying multiple Absolute Learning modules does not further improve the accuracy. We expect that attributes are a clean form of labels in contrast to {\em word2vec} and very complementary to class labels {\em cls.}
}

\section{Remaining experimental details.} For augmentations, we randomly apply resized crop (scale 0.6--1.0, ratio 0.75--1.33), horizontal+vertical flips, rotations ($0$--$360^{\circ}$), and color jitter. 
Annotated per class attribute vectors ({\em mini}Imagenet) have 31 attributes (5 environments, 10 colors, 7 shapes, 9 materials). For augmentation keys, taking rotation as example, we set a 4-bit degree to annotate random rotations, '0001' refers to rotations with $0\sim90^{\circ}$, '0010' refers to rotations with $90\sim180^{\circ}$.  

\end{appendices}

{\small
\bibliographystyle{ieee_fullname}
\bibliography{fsl}
}

\end{document}